\documentclass[letterpaper]{article} 
\usepackage{aaai25}  
\usepackage{times}  
\usepackage{helvet}  
\usepackage{courier}  
\usepackage[hyphens]{url}  
\usepackage{graphicx} 
\usepackage{natbib}  
\usepackage{caption} 
\usepackage {amsmath}
\usepackage{subfigure}
\usepackage{booktabs}
\usepackage{xcolor}
\usepackage{amsmath}
\usepackage{tabularx}
\usepackage{bigstrut}
\usepackage{adjustbox}
\usepackage{multirow}
\usepackage{makecell}
\usepackage{algorithm}
\usepackage{algorithmic}

\usepackage{newfloat}
\usepackage{listings}
\DeclareCaptionStyle{ruled}{labelfont=normalfont,labelsep=colon,strut=off} 
\floatstyle{ruled}
\newfloat{listing}{tb}{lst}{}
\floatname{listing}{Listing}
\title{Disentangle Nighttime Lens Flares{:} Self-supervised Generation-based Lens Flare Removal}

\author{
    Yuwen He\textsuperscript{\rm 1},
    Wei Wang\textsuperscript{\rm 1}\thanks{Corresponding author: wangwei8@wust.edu.cn},
    Wanyu Wu\textsuperscript{\rm 1},
    Kui Jiang\textsuperscript{\rm 2}
}
\affiliations{
    \textsuperscript{\rm 1}Computer Science and Technology and Hubei Province Key Laboratory of Intelligent Information Processing and Real-time Industrial System, Wuhan University of Science and Technology\\
    \textsuperscript{\rm 2}School of Computer Science and Technology, Harbin Institute of Technology\\
    heyuwen@wust.edu.cn, wangwei8@wust.edu.cn, wuwy0510@163.com, jiangkui@hit.edu.cn

}

\usepackage{bibentry}

\begin{document}

\maketitle

\begin{abstract}
Lens flares arise from light reflection and refraction within sensor arrays, whose diverse types include glow, veiling glare, reflective flare and so on. Existing methods are specialized for one specific type only, and overlook the simultaneous occurrence of multiple typed lens flares, which is common in the real-world, e.g. coexistence of glow and displacement reflections from the same light source. These co-occurring lens flares cannot be effectively resolved by the simple combination of individual flare removal methods, since these coexisting flares originates from the same light source and are generated simultaneously within the same sensor array, exhibit a complex interdependence rather than simple additive relation. To model this interdependent flares’ relationship, our Nighttime Lens Flare Formation model is the first attempt to learn the intrinsic physical relationship between flares on the imaging plane. Building on this physical model, we introduce a solution to this joint flare removal task named \textbf{S}elf-supervised \textbf{G}eneration-based \textbf{L}ens \textbf{F}lare \textbf{R}emoval \textbf{N}etwork (SGLFR-Net), which is self-supervised without pre-training. Specifically, the nighttime glow is detangled in PSF Rendering Network(PSFR-Net) based on PSF Rendering Prior, while the reflective flare is modelled in Texture Prior Based Reflection Flare Removal Network (TPRR-Net). Empirical evaluations demonstrate the effectiveness of the proposed method in both joint and individual glare removal tasks.
\end{abstract}
%

\section{Introduction}
Lens flares are common optical artifacts in nighttime photographs, which occur around light sources in images due to the scattering and reflection of light rays in the medium and reflection within the camera lens. It can be challenging to discern details surrounding the light sources as well as remove the unwanted reflective flare simultaneously, further hindering the downstream applications.

To tackle lens flares, most hardware solutions focus on improving the camera's optics to better eliminate and control flares \cite{boynton2003liquid,raskar2008glare,macleod2010thin,chen2010antireflection}. 
Inspired by the nighttime haze model including scattered airlight proposed by \cite{23}, \cite{park2016nighttime}, \cite{yang2018superpixel} and \cite{10} adopted nighttime dehaze methods to glow suppression tasks. Based on the smoothness of the glow, \cite{11} introduced an unsupervised light effect suppression method for clean weather.
To address individual reflective flares, the Flare7K \cite{5} stimulated the first reflective/ghost flares datasets in a synthetic manner but limited in Reflective flare patterns. Then, \cite{4} proposed a larger synthetic ghost flare database with different flare shapes for end-to-end training at nighttime.

\begin{figure}[!t]
\centering
\hspace{-5mm}
\includegraphics[width=0.49\textwidth]{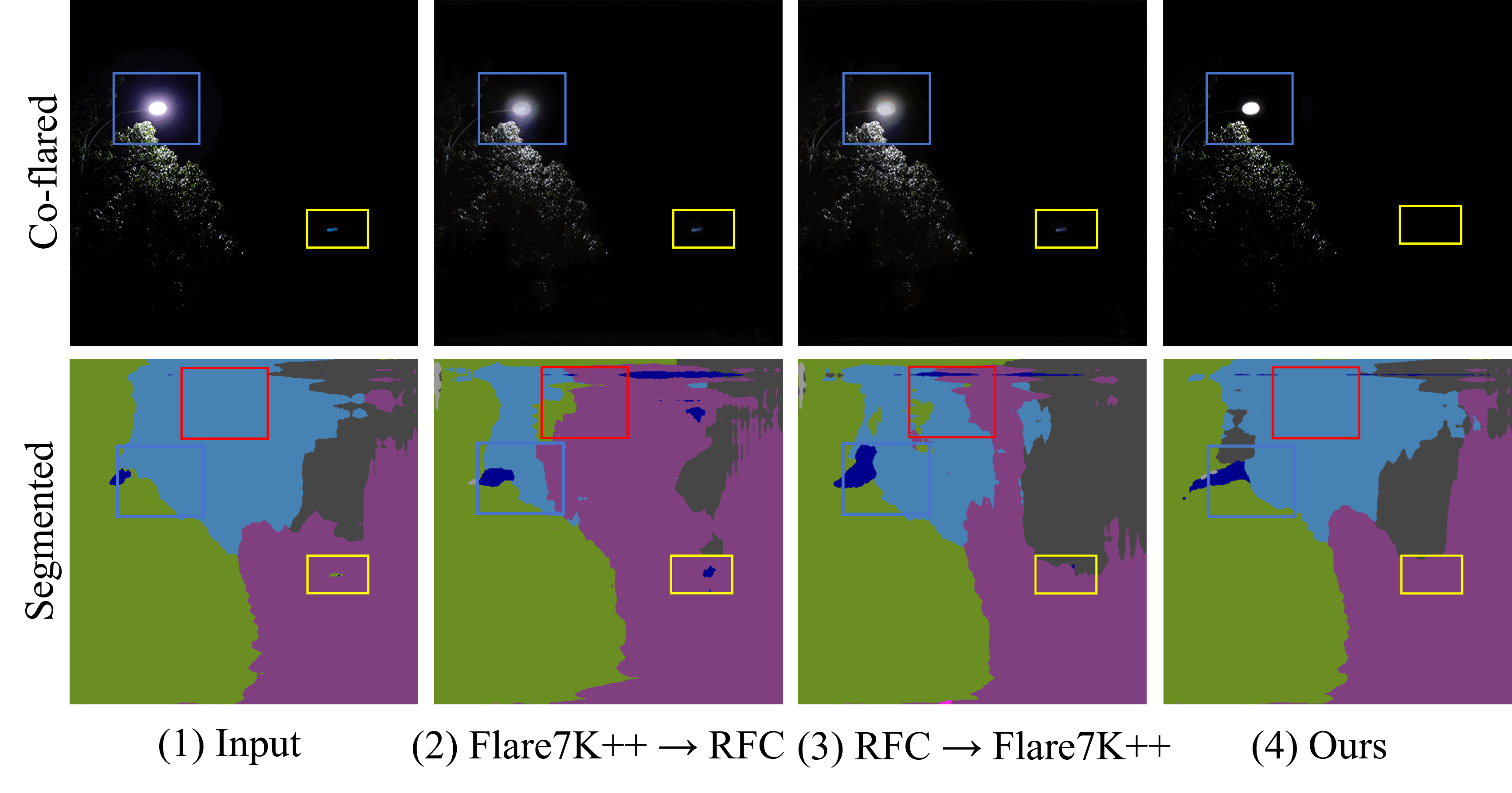}
\caption{ Up to down: (1) Coexisting flared input and its segmentation GT. (2) Deglowed via Flare7K++ \cite{20} and deghosted via RFC \cite{19} results with its segmented version via \cite{Wu_2021_CVPR}. (3) Deghosted and deglowed results with its segmented version. (4) Result and segmented result of our approach.}
\label{method}
\end{figure}

However, synthetic paired flare training datasets cannot accurately model the intricate imaging paths within lens arrays, especially with nighttime light sources. As an example shown in  Fig. \ref{method}, the flares co-existence cannot be easily solved via a two-stage cascaded solution in $(2)-(3)$, which is observed with uncleared glow (blue box) and  ghost (yellow box). These cascaded solution further lead to the misrepresentation of sky region (red box) in the semantic segmentation. Our method can solve the joint problem of glow and ghost and improve the results of semantic segmentation.\\
To explore their tangle relationship and offer a solution for this joint flare task, our approach firstly introduces a comprehensive Nighttime Lens Flare Formation Model which physically describe the joint formation of glow and reflective flares with two prior-based observation. Building upon this physical model, we propose a Generation-based Lens Flare Removal Network (SGLFR-Net), specifically designed to mitigate lens flare caused by artificial light sources at night, in a self-supervised manner and without pre-training or training data. 
Our SGLFR-Net employs a two-stream network, in which the PSF rendering network (PSFR-Net) stream stimulate the glow formation with PSF rendering prior, while the Texture Prior based Reflection Removal Network (TPRR-Net) stream produces an updating ghost version incorporated with PSFR-Net outcome, to be constrained by Optical Symmetry based Texture Prior Module (OS-TPM). In all, our contributions are as three-folds:
\begin{itemize}
\item Propose Nighttime Lens Flare Formation model for nighttime co-existing flares originated from same light source via optical paths to imaging plane simultaneously.
\item Derived from this physical model, our Self-supervised Generation-based Lens Flare Removal Network (SGLFR-Net) is introduced to form the first joint flare removal solution without pre-training or training data.
\item Our approach is validated effective in both joint flare removal and individual flare removal tasks in terms of both synthetic and real-world datasets.
\end{itemize}
\section{Related Works}
\noindent\textbf{Glow Removal. }
Driven by the nighttime flare image dataset Flare7K \cite{5}, several flare removal methods are based on supervised training \cite{6,7,15}. Flare7K, the first synthetic dataset for flare removal, has limitations in generality and type diversity, and the entanglement of light and noise \cite{guo2023low} complicates flare removal process. 
Given real-world pairing challenges of flare datasets, unsupervised training is viable.
\cite{11} integrates layer decomposition and suppression in a network, using the estimated light-effects layer to guide an unsupervised glow suppression network. Image dehazing is challenging \cite{wei11}, \cite{lin2024nighthaze} takes atmospheric light from a real haze image and renders it into a clear image, thereby suppressing the glow effect in self-supervised learning. \cite{Cong_2024_CVPR} employs a retraining strategy and semi-supervised training based on localized luminance windows to suppress the glow effect. APSF was first introduced to computer vision by \cite{31} with a physical imaging model under bad weather \cite{li21}. \cite{13,13k} used a zero-shot approach to generate APSF and transfer the task of glow suppression to the glow generation learning task, which solved the challenge of uneven glow. APSF is also generally applied in dehazing \cite{defogging29,zhang2017fast28,2,d3,30,10}. 
\textbf{However}, APSF is specifically derived to account for multiple scattering in the atmosphere and does not address glares from lens refraction inside the camera system during nighttime photography, which includes glow and "ghosting" effects.

\noindent\textbf{Reflection Removal. }
Specular reflection and lens flare are main concerns for image reflection. 
Data-driven reflection removal methods \cite{19,37,915,36} have become progressively more popular as the first reflection removal dataset for transmitted images \cite{35}.
Earlier studies \cite{1,12,chabert2015automated} on removing lens highlight flare formed by lenses often adopt a two-phase strategy, involving detection followed by removal. This approach often limits recognition to specific shapes, such as a "bright spot". \cite{34} proposed an alternative method using deconvolution of the measured flare diffusion function to remove lens flare, mainly removal rounded flares.  
For nighttime reflective flares, \cite{4} proposed an optical symmetry rule and developed a synthetic dataset for end-to-end training. \textbf{However}, the above studies addresses the physical formation of various types of lens flare induced by artificial night lighting nor proposes any solutions.
\begin{figure}[t]
\centering
\includegraphics[width=0.47\textwidth]{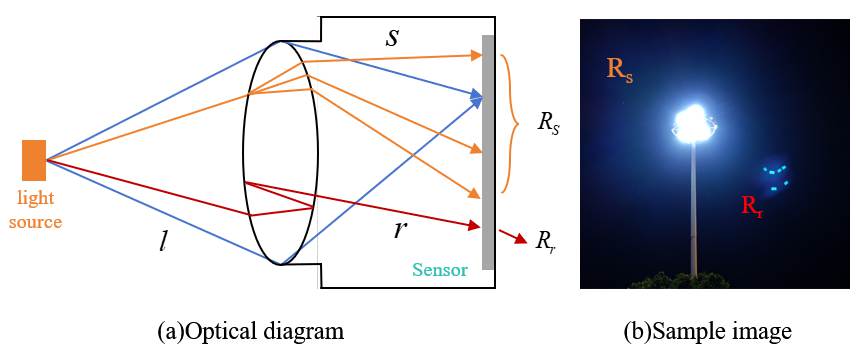}
\caption{(a) Optical path illustration from the light source to the imaging plane, where the blue line is the ideal light path. The orange line $s$ is the path of scattered light that produces glow flare $R_{s}$, and the red is the path of light refraction between lenses that produce the reflective flare (ghost) $R_{r}$, and $l$ is the line of incident light. (b) Example of lens glare.}
\label{op}
\end{figure}
\section{Proposed Physical Model}
\begin{figure*}[t] 
\centering 
\includegraphics[width=0.8\textwidth]{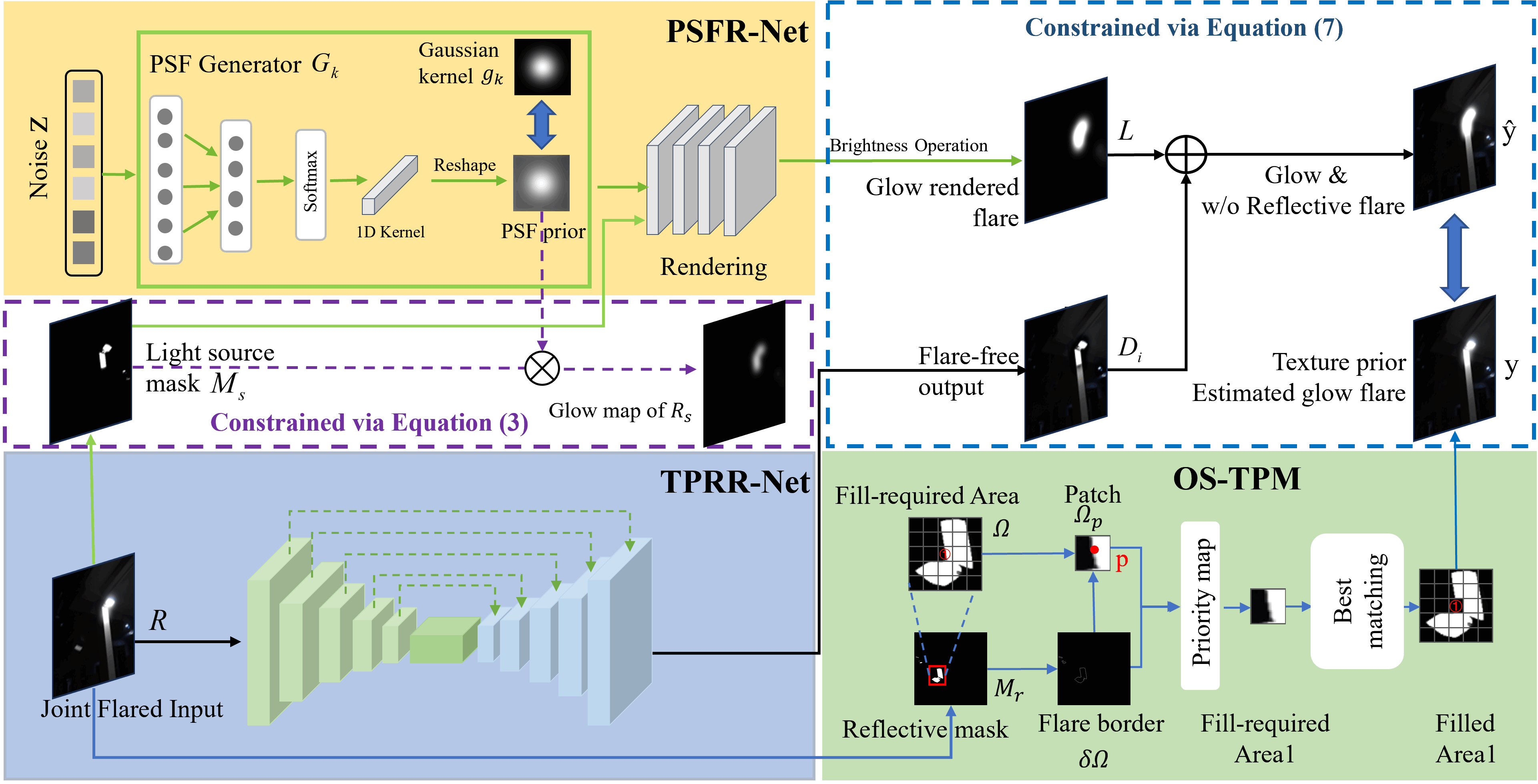} 
\caption{The proposed Self-supervised Generation-based Lens Flare Removal Network (SGLFR-Net) without pre-training is composed with PSF rendering network (PSFR-Net) and texture prior based reflection removal network (TPRR-Net), restricted with Optical Symmetry Based Texture Prior Module (OS-TPM). Two derived priors, namely PSF rendering prior and optical symmetry texture prior are incorporated into our SGLFR-Net on the basis of the Nighttime Lens Flare Formation model.} 
\label{network} 
\end{figure*}
\subsection{Nighttime Lens Flare Formation}
The formation process of nighttime lens flare (shown in Fig. \ref{op}) can be uniformly described as a transformation from the flare-free image $I$ to the lens-flared image $R$. This transformation is composed of the 2D lens flare spread functions $\hat{GTF}$ for each pixel in $I$, resulting in a 4D tensor $L(x, y, u, v)$. 
Then, this formation can be simplified with 2D convolution, when $\hat{GTF}$ is a shift-invariant function and made of all shift versions of $\operatorname{GTF}_{\mathbf{u}}$:
\begin{equation}
\small
\begin{aligned}
& ~~~~~~~~~~~~~~~~~~~~~~~~~R=I\ast\hat{GTF}, &&\\
& \operatorname{GTF}_{\mathbf{u}}(x)\!=\!\operatorname{GTF}_{\mathbf{v}}(x\!+\!(u-v)), \forall u, v\! \in \!1 \ldots n m. &&
\end{aligned}
\label{gtf}
\end{equation}
\noindent For simplicity, the 4D tensor is simplified into $nm\times 1$ vectors $I$ and $R$ with $nm\times nm$ transformation matrix.\\
\noindent\textbf{Glow Flare Physical Formation. }
The phenomenon of radiative diffusion of a light source is formed by multiple scattered lights within the camera lens, and we use $GTF_s$ to denote the physical formation of the glowing flare:
\begin{equation}
\small
\operatorname{GTF}_{s}\!\!\left\{k_i(r)\!\right\}_{i=1}^m\!\!=\!\!(1\!-\!\alpha) \delta\!\left(r\!\!-\!\!r_i\right)\!+\!\alpha\!\left(c\left(r, r_i\right)\!\!+\!\!b\left(r, r_i\right)\right),
\end{equation}
in which $k=(1-\alpha) \delta+\alpha(b+c)
$. $r\in(x,y)$ is the pixel location and $r_i$ is the spreading pixel of $r$ in the neighbouring $m$ pixels. $c$ is the operator kernel to capture the light source  and $b$ only captures the glow region $m$. Here $(1-\alpha) \delta+\alpha c$ in $k$ times corresponds to a central peak with a small finite size, which is usually described with the point spread function $PSF$. Therefore, the obtained image $R_s$ with glow flare is with glow flare transfer function $GTF_s$  as follows:
\begin{equation}
\small
R_s=I\ast\hat{GTF_s}=\sum_i(I_i\times M_s) \otimes PSF_i,
\label{glow}
\end{equation}
here $I$ is the glow-free input and $I_i$ is the scattered portion of direct transmission of $I$. $M_s$ is a glow region mask to indicate the glow's spatial location $(x,y)$ in $R_s$ with a widely used multiple scattered $GTF$.\\
\noindent\textbf{Reflective Flare Formation. } Reflection flare is caused by unknown cyclic reflections occurring between lens elements within a camera lens array, which presents a formidable obstacle in precisely capturing and modeling the intricate pathways of reflection flare, which is impossible to be accurately modelled for path uncertainty. Inspired by ray tracing, a light-ray-sampling method is employed.\\
When the incident ray $l$ undergoes reflection and refraction at the interface between two media, such as a lens and another lens. The refractive indices of the respective media are denoted as $n1$ and $n2$. The reflected ray is denoted as $r$. The normal vector $N$ of the interface lies in the same plane. The formula for obtaining the refracted ray $r$ is presented as:
\begin{equation}
\small
\boldsymbol{r}=\frac{n_1}{n_2} \boldsymbol{l}\!-\!\left(\sqrt{1\!-\!\frac{n_1^2}{n_2^2}\left(1\!-\!(\boldsymbol{l} \cdot \boldsymbol{N})^2\right)}+\frac{n_1}{n_2} \boldsymbol{l} \cdot \boldsymbol{N}\right) \boldsymbol{N},
\end{equation}
The law of reflection can be considered as a special case of the law of refraction, with the condition that $n2 = -n1$. The reflected ray $R_r$ to form a reflective flare can be expressed as follows:
\begin{equation}
\small
R_r= I\ast\hat{GTF_r}=\sum_j(I_j\times M_r) \otimes(2 (\boldsymbol{N})^2-1), 
\end{equation}
in which the $j$ portion of incident ray $l$ that reach the imaging plane in the reflective flare region $M_r$.\\
\noindent\textbf{Joint Flare  Formation. }
The incident ray $l$($GTF$) from the light source reaches the imaging plane with three types: the ideal light $I_0$ to form normal imaging, the scattered light that produces the glowing flare (scattering-GTF $GTF_s$), and the reflective light that produces reflective flare (reflection-GRSF $GTF_r$). By parameterizing the light field $L$ on the sensor, where $(x,y)$ represents the sensor plane and $(u,v)$ denotes the main lens of the aperture plane, the internal light field $f$ of the camera that reaches imaging plane is: 
\begin{equation}
\small
f= i_0 * \alpha(\delta+GTF_s+GTF_r),
\end{equation}
here $i_0$ is the ideal light. $\alpha$ is the normalization constant and $\delta$ controls the minmum ray reaches the imaging plane. Following this formula, the output image with glow and ghost flare $R$ can be expressed as:
\begin{equation}
\small
\begin{split}
R\!\!=\!\!I_0\!+\!I\!\!\ast\hat{GTF_s}\!+\!I\ast\hat{GTF_r}\\
\!\!=\!\!I_0 \!+\! G(R,M_s) \!+\! D(R),
\label{join}
\end{split}
\end{equation}
here $I_0$ is the ideal light of $R$ formed via a small portion of the intrinsic light, which is close to the flare-free input $I$ outside both glow and ghost flared region only. $G(R, M_s)$ is the estimated glow, which is excited by the proposed PSF Rendering Network (PSFR-Net). on the basis of $PSF$ modeling and the light source map $M_s$ obtained by weighting through R. $D(R)$ is obtained by the Texture Prior Reflection Flare Removal Network (TPRR-Net) by performing operations such as threshold segmentation and center rotation on $R$ via the Optical Symmetry-based Texture Prior Module (OS-TPM).
\section{Proposed Network}
\subsection{Self-supervised Generation-based Lens Flare Removal Network (SGLFR-Net)}
Fig. \ref{network} shows our pipeline for our Self-Supervised Generation of  Lens Flare Removal Network (SGLFR-Net), which consists of two streams, the PSF Rendering Network (PSFR-Net) based on the PSF rendering prior, and the Texture Prior Reflection Removal Network (TPRR-Net) based on the optical symmetry prior. Both prior are spatially related based on our nighttime lens flare formation and modelled uniformly.\\
\noindent\textbf{PSF Rendering Prior.}
According to widely-used light rendering approach \cite{tog05}, the light source passes through the scattering medium $T_{sv}$ to reach the inside of the lens, where it undergoes a scattering process $PSF(k)$ to form a glow. $PSF$ can serve as multi-scattered $GTF$. Notably, $PSF(k)$ is independent of the ideal image $I_0$ and the flared input $R$, then it is modeled individually and used as a guide for PSF Rendering Network.\\
\noindent\textbf{PSF Rendering Network(PSFR-Net).}
Different from generative networks \cite{17}, we use a simple but effective FCN network $G_k$ to compose a fuzzy kernel $k$, followed by a four-layer CNN convolved with a light source mask to simulate the glow. Then the proposed PSF rendering prior is introduced with a random noise $z$, sampled from a uniform distribution $[0,1]$. Then SoftMax nonlinearity is adopted to produce $1D$ output of $G_k$ and reshaped into a 2D fuzzy kernel $k$. To ensure this stimulated glow $B_{l}$ is similar to input $R$ in brightness, our Brightness Operation layer is formed as:\\
\begin{equation}
\small
L=B_l\times Ad_\sigma\times Ad_\varphi\times Ad_\beta,
\end{equation}
\begin{equation}
\small
Ad_\sigma=\mu\cdot Bri_{perc}(M_s)-\eta\cdot Bri_{perc}(M_s)+\nu, 
\label{ad1} \end{equation}
\begin{equation}
\small
Ad_\varphi\!\!=\!\!\begin{cases}\frac{Bri_{glob}(R)}{Bri_{glob}(B_{l1})},Bri_{glob}(B_{l1})\!\!<\!\!Bri_{glob}(R)\\1,\quad other\end{cases} ,
\label{ad2}
\end{equation}
\begin{equation}
\small
Ad_{\beta}=\frac{Bri_{\max}-Bri_{loc}(1-M_{s})}{Bri_{\min}} ,
\label{ad3}\end{equation}
where $Bri_{glob}$ is the global brightness operation. $B_{l1}$ is the glow map of $B_{l}$ adjusted by Eq. \ref{ad1}. $B_{l2}$ is the glow map adjusted by Eq. \ref{ad2}. $Bri_{loc}$ is the local brightness operation and $Bri_{\min}$ denotes the minimum between $B_{l2}$ and R.
\begin{figure}[t]
\centering
\includegraphics[width=0.44\textwidth]{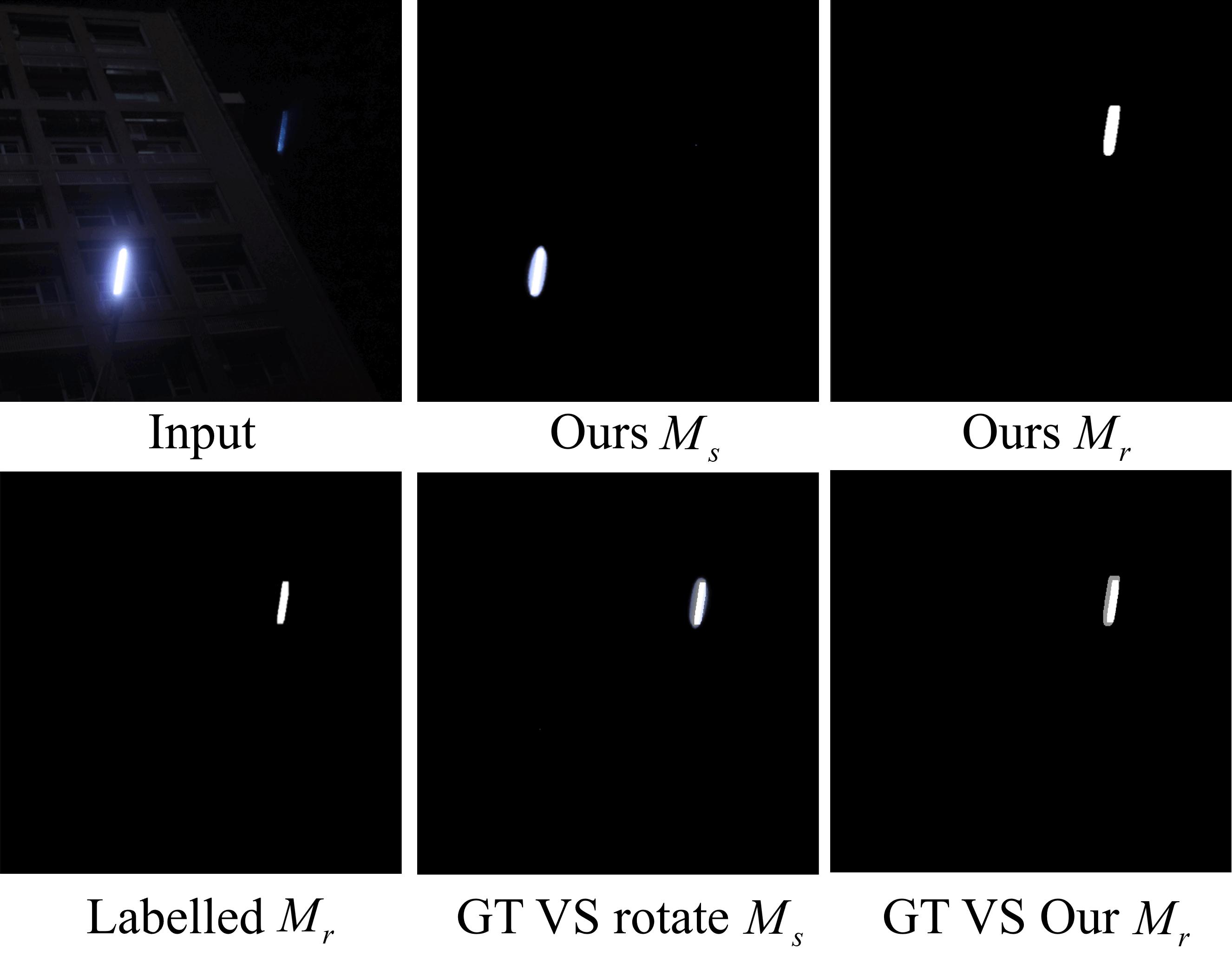}
\caption{Our Optical symmetry-based texture prior(OS-TPM) illustration. Light source mask $M_{s}$ is spatially related to $M_{r}$, obtained via OS-TPM module. Our $M_{r}$ completely covers the manually labelled GT $M_{r}$.}
\label{optical}
\end{figure}
\begin{figure*}[t]
\centering
\includegraphics[width=1\textwidth]{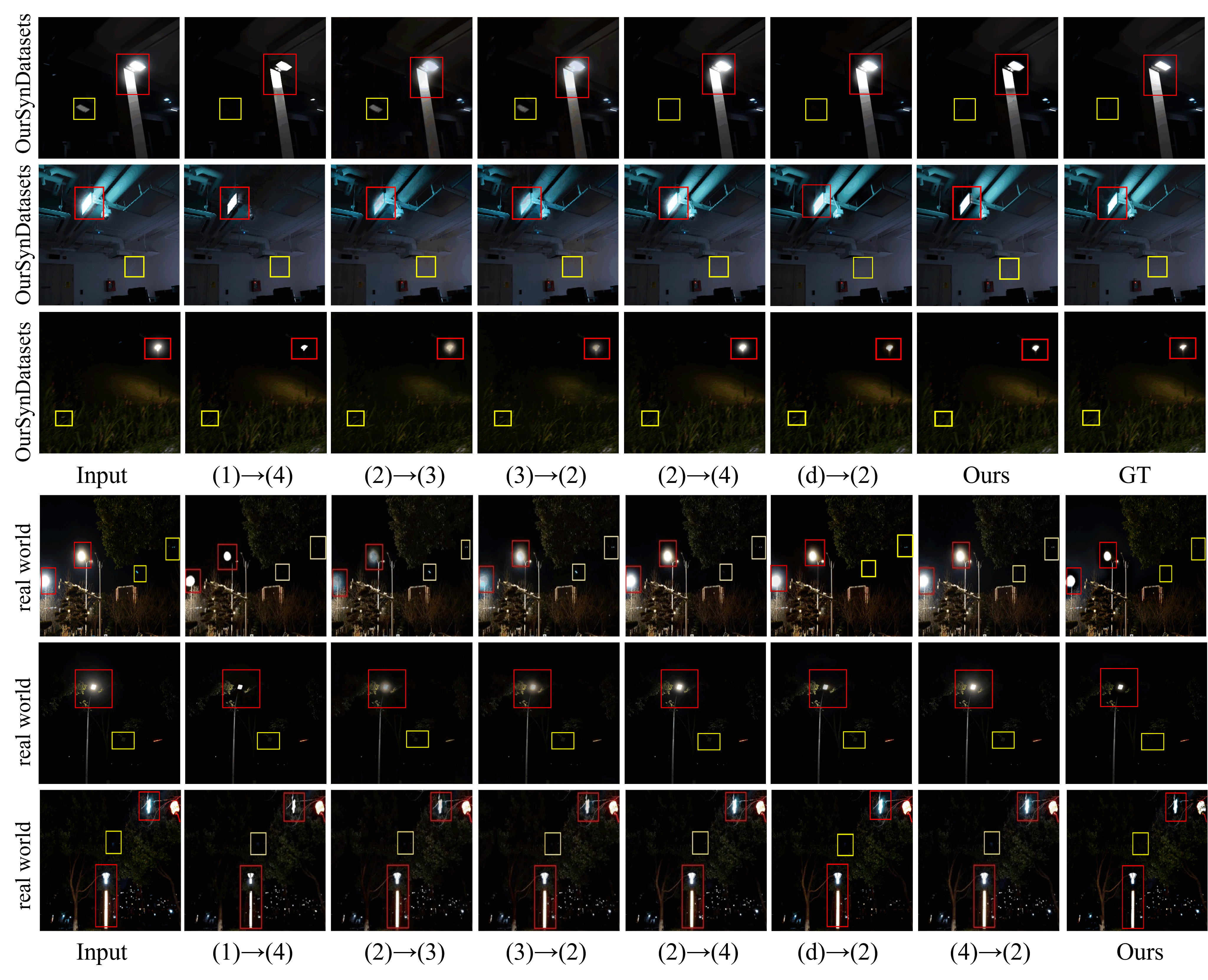}
\caption{\textbf{Joint glow and reflective flare removal Task}. Visual comparison of our SGLFR-Net and glow and reflection flare removal SOTA combinations in OurSynDatasets and real world datasets. Notation: (method1)$\rightarrow$(method2).}
\label{SynDatasets}
\end{figure*}
\\
\noindent\textbf{Optical Symmetry Based Texture Prior. }
The nighttime reflective flare is caused by the bright light source and this ghost obeys the optical symmetry rule when it is connected to the light source in a square image. In other words, Rotating the reflective flare mask GT $M_r$ around the optical center will result in its alignment with the light source mask $M_s$. Therefore, the glow flare $R_s$ and reflective flare $R_r$ are both a derivative version of nighttime lens flare formation in Eq. (\ref{gtf}), while they are spatially connected. This prior is validated in Fig. \ref{optical}, our proposed reflection flare map $M_r$ fully covers the reflection flare regions of the input $R$. Notably, our approach is self supervised without training. \\
In the following, we consider the elimination of reflection flares as a kind of in-texture mapping \cite{inpaint}, which is initialized by the Optical Symmetry-based Texture Prior (OS-TPM) module $y$ to guide the training of TPRR-Net.
In this process, $M_r$ is segmented and rotated $R$, and then detected for the contour boundary $\delta \Omega$ of $M_r$,  whose pixels $p$ are labeled $\mathbf{p} \in \delta \Omega$. Given a patch $\Omega_p$ centred at the pixel $p$, its refined order is defined with priority map:
\begin{equation}
\small
P(\mathbf{p})=C(\mathbf{p}) T(\mathbf{p}),
\end{equation}
in which $C(\mathbf{p})=\frac{\sum_{\Omega p(i,j)}confidence}{area(\Omega_p)}$ stores the maximum confidence score in flared region $\Omega$.
$T(\mathbf{p})=\frac{|\nabla R_{\mathbf{p}}\cdot\mathbf{n_{p}}|}{\alpha}$ is the data term that incorporates reliable pixels around $p$, stands for the filling strength on border $\delta \Omega$. $\nabla R_{\mathbf{p}}$ is the isophote of $p$, $\mathbf{n}_{\mathbf{p}}$ is the unit vector to border $\delta \Omega$ at $p$. After the patch $\Omega_p$  has been filled, the confidence $C(\mathbf{p})$ is updated  via $C(\mathbf{p})=C(\hat{\mathbf{p}}) \quad \forall \mathbf{p} \in\Psi_{\hat{\mathbf{p}}} \cap \Omega$. Therefore, the highest priority contour point pairs are searched in OS-TPM and the best match operation is used to compute the source patch $\Phi_{\mathrm{s}}$ that best matches $\Omega_{p}$ to fill and form the desired $y$ via:
\begin{equation} 
\small
diff=\sum(\Phi_\mathrm{s}-\Omega_\mathrm{p})^2+\rho+\left|\nabla S_\mathrm{p}-\nabla R_\mathrm{p}\right|-\cos(\theta).\end{equation}
where $\rho$  is the Eulerian distance between the $\Phi_\mathrm{s}$ center point $s$ and the $\Omega_\mathrm{p}$ center point $p$, $\nabla S_\mathrm{p}$ is the $s$ isophote intensity, $\nabla R_\mathrm{p}$ is the $p$ isophote intensity.
\begin{figure*}[h]
\centering
\includegraphics[width=1\textwidth]{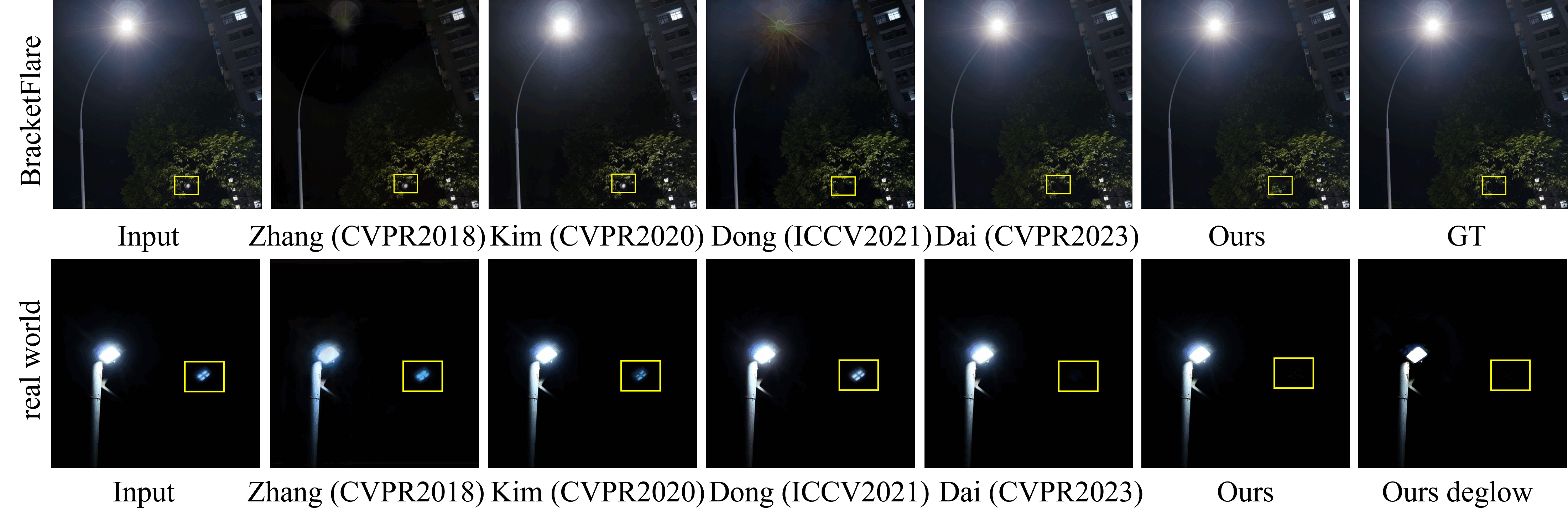}
\caption{Reflective Flare Removal Task. Visual comparison of individual reflection removal task with SOTAs on BracketFlare Datasets and real world captured images.}
\label{Reflective}
\end{figure*}
\begin{figure*}[h]
\centering
\includegraphics[width=1\textwidth]{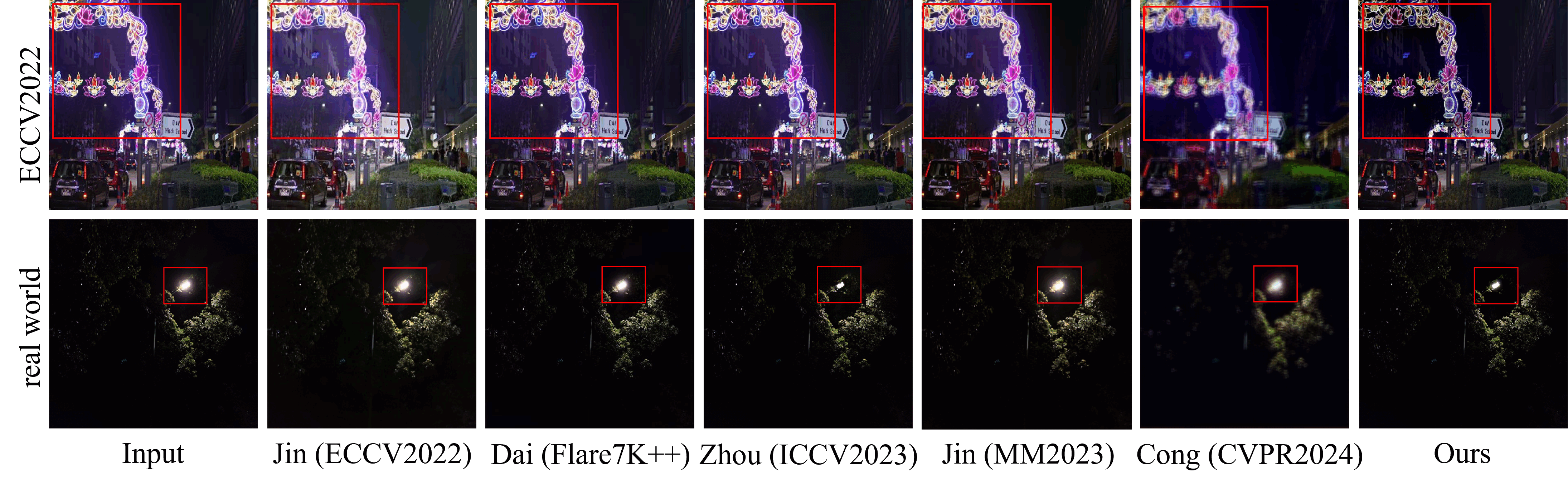}
\caption{Glow Suppression Task. Our visual comparison on individual glow suppression task with SOTAs on ECCV2022 datasets and real world captured nighttime glows.}
\label{glow}
\end{figure*}
\begin{table}[h]
  \centering
  \scalebox{0.74}{
    \begin{tabular}{c|c|c|c|c}
    \hline
    Methods & Strategy & Pipeline & \multicolumn{1}{c|}{PSNR↑} & \multicolumn{1}{c}{SSIM↑} \bigstrut\\
    \hline
    Glow Removal & \multicolumn{4}{c}{Deglow → Deghost} \bigstrut\\
    \hline
    \multirow{2}[2]{*}{(1) ICCV2023} & \multirow{2}[2]{*}{Supervised} & (1)→(3) & 27.76 & 0.962 \bigstrut[t]\\
    \multicolumn{1}{c|}{} & \multicolumn{1}{c|}{} & (1)→(4) & \textcolor{blue}{29.23} & 0.959 \bigstrut[b]\\
    \hline
    \multirow{2}[2]{*}{(2) Flare7K+ +} & \multirow{2}[2]{*}{Supervised} & (2)→(3) & 28.58 & 0.969 \bigstrut[t]\\
    \multicolumn{1}{c|}{} & \multicolumn{1}{c|}{} & (2)→(4) & 29.21 & 0.965 \bigstrut[b]\\
    \hline
    \multirow{2}[2]{*}{(a)ACM MM2023 } & \multirow{2}[2]{*}{Semi-Supervised} & (a)→(3) & 25.31 & 0.892 \bigstrut[t]\\
    \multicolumn{1}{c|}{} & \multicolumn{1}{c|}{} & (a)→(4) & 25.67 & 0.899 \bigstrut[b]\\
    \hline
    \multirow{2}[2]{*}{(b) ECCV2022} & \multirow{2}[2]{*}{Unsupervised} & (b)→(3) & 26.63 & 0.945 \bigstrut[t]\\
    \multicolumn{1}{c|}{} & \multicolumn{1}{c|}{} & (b)→(4) & 26.77 & 0.940 \bigstrut[b]\\
    \hline
    \multirow{2}[4]{*}{(e) CVPR2024} & \multirow{2}[4]{*}{Semi-Supervised} & (e)→(3) & 22.34 & 0.784 \bigstrut\\
\cline{4-5}    \multicolumn{1}{c|}{} & \multicolumn{1}{c|}{} & (e)→(4) & 24.57 & 0.826 \bigstrut\\
    \hline
    Reflective Flare Removal & \multicolumn{4}{c}{Deghost → Deglow} \bigstrut\\
    \hline
    \multirow{2}[2]{*}{(3) CVPR2018} & \multirow{2}[2]{*}{Supervised} & (3)→(1) & 27.41 & 0.964 \bigstrut[t]\\
    \multicolumn{1}{c|}{} & \multicolumn{1}{c|}{} & (3)→(2) & 28.71 & 0.969 \bigstrut[b]\\
    \hline
    \multirow{2}[2]{*}{(4) CVPR2023} & \multirow{2}[2]{*}{Supervised} & (4)→(1) & 28.87 & 0.963 \bigstrut[t]\\
    \multicolumn{1}{c|}{} & \multicolumn{1}{c|}{} & (4)→(2) & 28.83 & 0.966 \bigstrut[b]\\
    \hline
    \multirow{4}[4]{*}{(c) CVPR2020} & \multirow{4}[4]{*}{Supervised} & (c)→(1) & 26.83 & 0.939 \bigstrut[t]\\
    \multicolumn{1}{c|}{} & \multicolumn{1}{c|}{} & (c)→(2) & 26.34 & 0.948 \bigstrut[b]\\
\cline{3-5}    \multicolumn{1}{c|}{} & \multicolumn{1}{c|}{} & (c)→(a) & 24.62 & 0.867 \bigstrut[t]\\
    \multicolumn{1}{c|}{} & \multicolumn{1}{c|}{} & (c)→(b) & 25.58 & 0.927 \bigstrut[b]\\
    \hline
    \multirow{4}[4]{*}{(d) ICCV2021} & \multirow{4}[4]{*}{Supervised} & (d)→(1) & 27.73 & 0.961 \bigstrut[t]\\
    \multicolumn{1}{c|}{} & \multicolumn{1}{c|}{} & (d)→(2) & 28.93 & \textcolor{blue}{0.970} \bigstrut[b]\\
\cline{3-5}    \multicolumn{1}{c|}{} & \multicolumn{1}{c|}{} & (d)→(a) & 26.10 & 0.898 \bigstrut[t]\\
    \multicolumn{1}{c|}{} & \multicolumn{1}{c|}{} & (d)→(b) & 27.23 & 0.943 \bigstrut[b]\\
    \hline
    \multicolumn{5}{c}{Joint task model} \bigstrut\\
    \hline
    \textcolor[rgb]{ 1,  0,  0}{Ours} & \textcolor[rgb]{ 1,  0,  0}{Self-supervised} & \textcolor[rgb]{ 1,  0,  0}{Ours} & \textcolor[rgb]{ 1,  0,  0}{29.65} & \textcolor[rgb]{ 1,  0,  0}{0.974} \bigstrut\\
    \hline
    \end{tabular}%
    }
    \caption{Results on the SynDatasets for joint glow and reflective/ghost flares removal task. The best results are marked in \textcolor{red}{red}, and the second-best results are in \textcolor{blue}{blue}.}
  \label{Union_metrics}%
  
\end{table}%

\noindent\textbf{Texture Prior Based Reflection Flare Removal Network (TPRR-Net).}
Our TPRR-Net uses an Unet-like encoder/decoder architecture with a 5-layer encoder and decoder, as shown in Fig.\ref{network}. The flare-damaged image $R$ is hopped through the TPRR-Net to produce a glow and ghost flare-free output $D_i$ with $i$ epochs. This $D_i$ is self-supervised by $y$ with its incorporated ghost-free and glowed result $\hat{y}$, which is introduced in combination with glow-rendered result $L$ from PSFR-Net. Since the suppression of the glow effect affects the recovery of the light source, we add the weighted light source map to the output $D_i$. We estimated $y$ without ghosting and glowed by OS-TPM, and guided the training of SGLFR-Net by $\mathcal{L}_{MSE}$ and  1-$\mathcal{L}_{SSIM}$ measures between $y$ and $\hat{y}$. Our SGLFR-Net is unpretrained with any training data and universally solves glow and reflective lens flares.
\section{Experimental Results}
\subsection{Experimental Details}
\noindent\textbf{Settings.}
Our method is implemented on Pytorch with an NVIDIA GPU (version RTX 3090) and covers $3000$ iterations. For the fuzzy kernel, we sampled z from a uniform distribution from 0 to 1, with a fixed random seed of $0$.\\
\noindent\textbf{Dataset.}
The joint task datasets include the self-synthesized OurSynDatasets joint glow and ghost flared dataset (with GTs) and real world captured dataset(without GT). OurSynDatasets is stimulated from images in training dataset from BracketFlare \cite{4} processed with a function $\gamma$$ \sim $$\text{U(1.4,1.8)}$, since there are no joint flared datasets available. The Reflective Flare Removal task dataset is validated on the \cite{4} benchmark and real world captured dataset. The glow suppression task is validated on ECCV2022 \cite{11}(without GT) light effect dataset and real world captured datasets.
\subsection{Joint Task}
Caparisons were conducted with a combination of the latest de-glow and de-ghost methods, since no one has studied this joint glow and reflective/ghost flare removal task.  Five glow removal methods used here are Flare7k++ \cite{20}, ICCV2023 \cite{21}, MM2023 \cite{10}, ECCV2022 \cite{11}, CVPR2024 \cite{Cong_2024_CVPR} (retrained on \cite{NightHazy} dataset). Four reflection removal methods used here are CVPR2018 \cite{19}, CVPR2020 \cite{16}, ICCV2021 \cite{18}, CVPR2023 \cite{4}. The combinations with relative high performance are reported here, and more joint task results with unreported combinations can be found in our supplementary material.\\
\noindent\textbf{Qualitative Evaluation.} From the third row of the Fig.\ref{SynDatasets}, it can be seen that the method proposed in this paper can recover the shape of the light source similar to GT. With the exception of the (1)$\rightarrow$(4) combination and our method, none of the other methods are effective in removing the joint flare problem. The (1)$\rightarrow$(4) combination in the real image of the third row of images is not effective in removing glow and ghosting, nor can it deal with reflective flare and blue glow, which can be removed by our method. In the second case of real world, only our approach totally removed the ghost in the yellow box, other SOTAs produced results with uncleared edges of the ghost.\\
\noindent\textbf{Quantitative Evaluation.} Table \ref{Union_metrics} illustrates our approach ranks first in both PSNR and SSIM evaluations in joint glow and reflective/ghost flares removal task, in terms of $1.4\%$ and $0.4\%$ advantages compared to SOTAs respectively. Notably, our self-supervised approach needs no training data, unlike supervised comparative methods. This demonstrates that our Nighttime Lens Flare Formation contributes to our leading performance in synthetic and real world datasets.
\subsection{Reflective Flare Removal Task}
It is observed in Fig. \ref{Reflective} that only CVPR2023 \cite{4} and our approach fully removes the ghost flares while  results of other SOTAs have uncleared ghosts. It is also validated in Table. \ref{Reflective_table} as our approach ranks second as the only self-supervised method. The first place \cite{4} is a fully-supervised approach training on the same database, while our approach does not use any training data and features in model generalization.
\subsection{Glow Suppression Task}
It is clearly observed from Fig. \ref{glow} that our model almost fully eliminates the glow arond the light source, such that the shape of the light source is reliably recovered. The ICCV 2023 also demonstrates good performance in small light source recovery in the second case, but cannot estimate the irregular shaped glow around the complex light in the first case. Meanwhile, results from ICCV 2023, MM2023, ECCV 2022, CVPR 2024, and Flare7K++ still show unsatisfying veiling flare in Fig. \ref{glow}.
\begin{table}[h]
  \centering
   \scalebox{0.725}{
    \begin{tabular}{c|c|c|c|c}
    \hline
     Methods & Strategy & \multicolumn{1}{c|} {Training Datasets} & \multicolumn{1}{c|}{SSIM↑} & LPIPS↓ \bigstrut\\
    \hline
    \makecell{Dai\\(CVPR2023)} & Supervised &  \makecell{BracketFlare\\440 pairs} & \textcolor{red}{0.994} & \textcolor{red}{0.004} \bigstrut\\
    \hline
    \makecell{Zhang \\(CVPR2018)} & Supervised & \makecell{Reflection\\5000 pairs} & 0.83  & 0.074 \bigstrut\\
    \hline
    \makecell{ Dong\\ (ICCV2021)} & Supervised & \makecell{synthetic reflection\\14240 pairs} & 0.907 & 0.041 \bigstrut\\
    \hline
     \makecell{Soomin Kim \\(CVPR2020)} & Supervised & \makecell{physically-synthesized\\ 5000 pairs} & 0.857 & 0.069 \bigstrut\\
    \hline
    Ours  & \textcolor{red}{Self-supervised} & \textcolor{red}{No training}  & \textcolor{blue}{0.967} & \textcolor{blue}{0.038} \bigstrut\\
    \hline
    \end{tabular}%
    }
     \caption{ Comparisons between our approach and reflection removal methods on BracketFlare dataset. }
  \label{Reflective_table}%
\end{table}%
\begin{table}[t]
  \centering
     \scalebox{0.8}{
    \begin{tabular}{c|c|c|c}
    \hline
    Network structure & W/O PSFR-Net  & W/O TPRR-Net  &  Ours \bigstrut\\
    \hline
    PSNR$\uparrow$  & 27.78  &    29.61   & \textcolor[rgb]{ 1,  0,  0}{29.65} \bigstrut\\
    SSIM$\uparrow$ & 0.973 &    0.972   & \textcolor[rgb]{ 1,  0,  0}{0.974} \bigstrut\\
    \hline
    Loss Function & W/O 1-$\mathcal{L}_{SSIM}$  & W/O $\mathcal{L}_{MSE}$  & Ours \bigstrut\\
    \hline
    PSNR$\uparrow$  & 28.18 & 29.37 & \textcolor[rgb]{ 1,  0,  0}{29.65} \bigstrut\\
    SSIM$\uparrow$ & 0.965 & 0.972 & \textcolor[rgb]{ 1,  0,  0}{0.974} \bigstrut\\
    \hline
    \end{tabular}%
    }
    \caption{Ablation studies of the proposed method.}
  \label{abla_table}%
\end{table}%
\begin{figure}[H]
\centering
\includegraphics[width=0.45\textwidth]{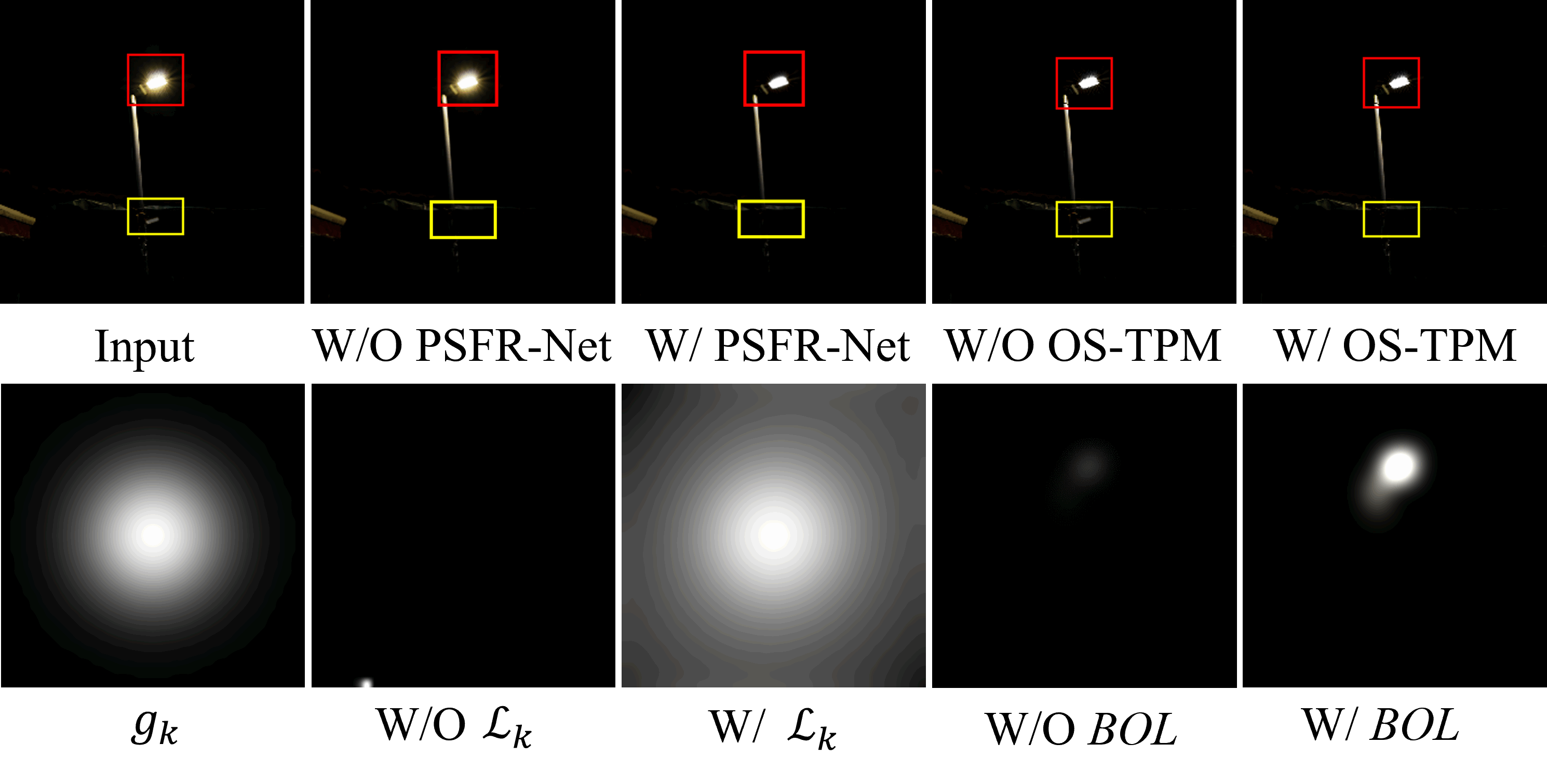}
\caption{Ablation studies of the proposed method.}
\label{abstudy}
\end{figure}
\subsection{Ablation Study}
\noindent\textbf{Loss function.}
Table. \ref{abla_table} have validated the effectiveness of $\mathcal{L}_{MSE}$ and 1-$\mathcal{L}_{SSIM}$ loss function of the proposed Self-supervised Generation-based Lens Flare Removal Network (SGLFR-Net), which is regarded as $\mathcal{L}_k$. This $\mathcal{L}_k$ is also used in the original the PSF prior generation, shown in Fig. \ref{abstudy} to validate its performance in shape regularization in PSF prior generation. In particularly, our model uses the $\mathcal{L}_{MSE}$ loss in the first $1000$ iterations to fit the pixel-level differences between $\hat{y}$ and $y$, $g_{k}$ and $k$. 1-$\mathcal{L}_{SSIM}$ loss is added in the latter $2000$ iterations to fit the differences in the spatial dimensions of the images. \\
\noindent\textbf{Network Structure.} 
Our two streams of the proposed SGLFR-Net are validated individually here. To verify the contribution of our TPRR-Net stream, we replaced it with ten-layer CNN architecture, such that it is observed a performance drop in Table. \ref{abla_table}. Then, the PSFR-Net stream is verified with its removal, which is observed with performance degradation in both index evaluation of Table. \ref{abla_table} and visual comparison in in Fig. \ref{abstudy}. This demonstrates the PSFR-Net contributes to locally light source shapes. More ablation experiments on OS-TPM is shown in Fig. \ref{abstudy}, which verifies OS-TPM contributes to the location of reflective/ghost flares. At last, our Brightness Operating Layer (BOL) has been verified to align the brightness of glow rendered image $L$ closer to the input $R$.
\section{Conclusions}
In this paper, the Nighttime Lens Flare Formation model is first introduced to give a uniform description of lens flares, such as glow flare and reflective flare (ghost). Based on this model, a PSF-rendering prior is formed to stimulate the glow formation independently and incorporated into the proposed PSF Rendering Network (PSFR-Net) stream. The other model-based prior, namely Optical Symmetry Based Texture Prior(OS-TP) is introduced to connect the spatial relation between glow and ghost, and used in the Optical Symmetry Based Texture Prior Module (OS-TPM) to guide the training of our whole Self-supervised Generation-based Lens Flare Removal Network (SGLFR-Net), in which no pre-training is needed. The other stream of our SGLFR-Net is our texture prior based reflection removal network (TPRR-Net), which produces an intermediate result with reflection and without glow and is incorporated with PSFR-Net outcome, to conduct a posteriori with OS-TPM estimation. Experiments have validated that our method is effective in both joint and individual lens flare tasks in both synthetic and real world dataset, without pre-training.

\section{Acknowledgments}
This work was supported financially by the Natural Science Foundation of China (Grant No.62202347) and the Open Research Fund from Guangdong Laboratory of Artificial Intelligence and Digital Economy (SZ) under Grant No.GML-KF-24-09.

\bibliography{references}

\end{document}